\PassOptionsToPackage{table}{xcolor}
\documentclass[acmlarge]{acmart}

\usepackage{tikz}
\usepackage{booktabs} % For formal tables
\usepackage{multirow}

\usepackage[ruled]{algorithm2e} % For algorithms

\SetAlFnt{\small}
\SetAlCapFnt{\small}
\SetAlCapNameFnt{\small}
\SetAlCapHSkip{0pt}
\IncMargin{-\parindent}

\usepackage{subcaption}
\usepackage{adjustbox}
\usepackage{slashbox}

\begin{document}
% Title portion
\title{Computational Attention System for Children, Adults and Elderly}
\author{Onkar Krishna}
\orcid{1234-5678-9012-3456}
\affiliation{%
  \institution{The University of Tokyo}
  \streetaddress{7-3-1 Hongo, Bunkyo-ku}
  \city{Tokyo}
  \state{Tokyo}
  \postcode{1138656}
  \country{Japan}}
\author{Kiyoharu Aizawa}
\affiliation{%
  \institution{The University of Tokyo}
  \city{Tokyo}
  \country{Japan}
}
\author{Go Irie} 
\affiliation{%
 \institution{NTT Corporation}}

\begin{abstract}
The existing computational visual attention systems have focused on the objective to basically simulate and understand the concept of visual attention system in adults. Consequently, the impact of observer's age in scene viewing behavior has rarely been considered. This study quantitatively analyzed the age-related differences in gaze landings during scene viewing for three different class of images: naturals, man-made, and fractals. Observer's of different age-group have shown different scene viewing tendencies independent to the class of the image viewed. Several interesting observations are drawn from the results. First, gaze landings for man-made dataset showed that whereas child observers focus more on the scene foreground, i.e., locations that are near, elderly observers tend to explore the scene background, i.e., locations farther in the scene. Considering this result a framework is proposed in this paper to quantitatively measure the depth bias tendency across age groups. Second, the quantitative analysis results showed that children exhibit the lowest exploratory behavior level but the highest central bias tendency among the age groups and across the different scene categories. Third, inter-individual similarity metrics reveal that an adult had significantly lower gaze consistency with children and elderly compared to other adults for all the scene categories. Finally, these analysis results were consequently leveraged to develop a more accurate age-adapted saliency model independent to the image type. The prediction accuracy suggests that our model fits better to the collected eye-gaze data of the observers belonging to different age groups than the existing models do. 
\end{abstract}



\maketitle

\renewcommand{\shortauthors}{K. Onkar et al.}

\section{Introduction}

The human visual system has the ability to filter out the most relevant part within the large amount of visual data, which is determined by the mechanism of selective attention of the human brain. In order to mimic this mechanism of human's selective attention in computational systems, researchers in computer vision have developed computational attention systems during last 15-20 years. These computational systems attempts to predict the most salienct location for the given input image similar to human observers. However, these systems \cite{am1, am2, am3, am4} fail to predict gaze accurately when compared with an actual human gaze. %as they are not tuned for inter-individual differences in scene viewing. 

The factors that drive a human's attention can be either related to the scene being observed or to the observer viewing the scene. In a sense, we can also think of fixation as either being `pulled' to a specific location by the visual properties of the fixated region i.e. scene related factors or `pushed' to a certain location by cognitive factors i.e. observer related factors such as task to be accomplished and the knowledge structure. Most of the research in developing computational attention system has been devoted to the scene related factors, as the bottom-up image features such as color, intensity and orientation are easier to model than the cognitive factors related to observers. However, there are a few studies where they investigated the role of observer related factors and also attempted to model them in computational system. The observer related factors are mainly studies for the cognitive factors such as given task  \cite{tatler2010yarbus, hayhoe2005eye}, human tendency \cite{am7}, habituation and conditioning \cite{am8}, and emotions \cite{am9}, while the impact of physical factors over scene viewing behavior which are predominately related to the external state of human observers, such as observer's age, eye sight, visual disparity, and gender are least explored. In this paper, we focus on the age-impact on fixation tendencies while viewing natural scenes. 

\begin{figure*}
\centering
  \includegraphics[width=\textwidth, height=6cm]{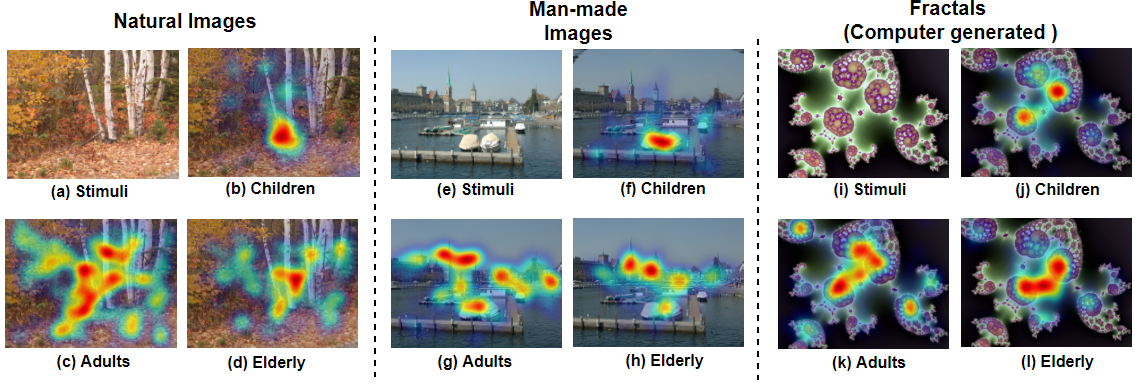}
  \caption{\textit{Sample stimulus from natural, man-made, and fractals categories and their heat map generated from the recorded eye-gaze data of children, adults and elderly participants.}}
  \label{fig:teaser}
\end{figure*}

\begin{figure*}
\centering
  \includegraphics[width=\textwidth, height=6cm]{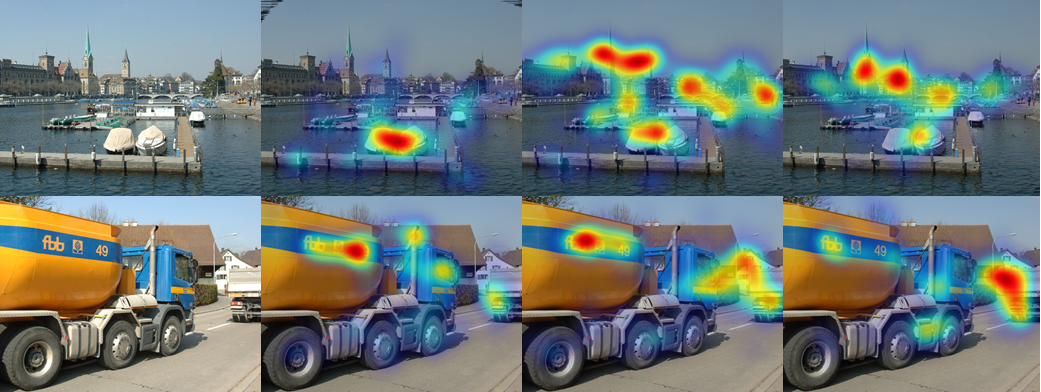}
  \caption{\textit{First column from left shows the example stimuli used in our study. The second, third, and fourth columns are heat maps representing salient locations attended by children (18 participants), adults (23 participants), and elderly (17 participants), respectively, in an eye-tracking experiment.}}
  \label{fig:teaser}
\end{figure*}

In recent years, vision research studies \cite{h, i} have shown that age-related changes in scene-viewing behavior exist by measuring the age-impact on gaze attributes, such as total number of fixations per image, inter-fixation distance, fixation duration, and image feature related viewing across different age groups. To date, however, %to the best of our knowledge no study that quantitatively analyzes the age-related changes in scene-viewing behavior in the context of development of an age-adapted computational attention system has been conducted.
most of the existing computational models predict salient locations pertaining to young adults \cite{am2, am4} and ignore the variations that comes with age. In this study, we discovered that observers belonging to the same age group and looking at the same scene have differences in gaze landings that become more prominent when the observers belong to different age groups, as shown in Figure 1, children, adults, and the elderly age groups have different spread of fixations while exploring the scene and this stands valid for different scene categories. Similarly, it can be seen in Figure 2 that children and elderly visualize different level of depth in a scene.      

In this study we have not only investigated the age-related differences scene viewing behavior but also the differences in gaze pattern across the images belonging to three different categories; "man-made", "nature", and "fractals". This investigation is triggered by the fact that human gaze is not only driven by visual input but also the knowledge structure associated with the input \cite{navalpakkam2006top}. It has been known for few decades that the gist of a scene can be grasped very rapidly, within the duration of only a fixation \cite{unema2005time}. Given the fast gist understanding, the category of the image such as natural, man-made, and fractals are known in very beginning of the scene viewing \cite{wu2014guidance}, later the prior knowledge about image category guide the gaze together with the bottom-up features associated with the specific image currently in view. Our main aim of including different scene categories was to analyze that age-related differences in scene viewing behavior is independent of the class of the scene in observation.   

In this paper, quantitative analysis of the age-related differences of fixation tendencies are performed by using four objective measures including depth bias, center bias, explorativeness, and inter-individual similarity. Consequently, the analysis results provided us some recommendations that we incorporated in the general framework of learning based saliency model to make it age-adapted. This involve choosing suitable scale of features in accordance with the level of detail in a scene to which observers of different age groups explored. And including differently weighted depth map as a new feature based on different tendencies of children and elderly toward fixating on foreground and background. Similarly inter-individual similarity and center bias are used in the proposed model to tune it further to observers age. We demonstrate that our new model yields better prediction accuracy compared with those which do not consider age-related factors. In this study, we observed the explicit tendency of foreground bias in children observers, and developed a framework to quantify this tendency.  

%Depth bias measures the percentage of fixations landed on the most salient foreground and background locations of the scene across age groups, whereas central bias measures age related differences in the tendency of fixating around the center while exploring the scene belonging to three different categories. Explorativeness evaluates the spread of the explored locations across age groups and image type, whereas the inter-individual similarity quantifies the agreement between explored regions within the same or different age groups. 

The following are the three main contributions of our study. First, we extended the study reported in \cite{am111,luna2008development} to exploration of the differences in the scene-viewing behavior of observers in three different age groups: children, adults, and the elderly during inspection of "natural", "man-made" and "fractal" stimulus. Second, we developed new parameters, depth bias, inter-individual similarity and upper performance limit to analyze the age impact on scene viewing behavior. Depth bias examined weather observers of different ages have different tendencies toward fixating on foreground and background. Finally, we incorporated the analysis results in the development of an age-adapted computational attention system.    
% Head 1
\section{Background}
We first review the available literature on developmental changes in scene viewing behavior and specify the developmental factors that missed in the existing studies that need to be quantified in order to develop an age-adapted saliency model. We then briefly review works on computational model of visual attention system for adult observers and discuss the scope of upgrading them to an age-adapted saliency model for complex image belonging to different scene categories. %We also review a very recent work on age-adapted saliency model for children and adult observers, and their drawbacks. 
     
\subsection{Developmental Literature}

In this section we present the evidences of age-impact on scene viewing behavior from developmental literature in order to establish the need for quantifying the age-impact on scene viewing to develop an age-adapted saliency model. There are many studies in psychology and neuroscience investigating the aging effect on eye-movement controls \cite{luna2008development, aring2007visual, irving2006horizontal, fukushima2006vestibular, klein2001developmental} such as focusing on target, maintaining focus on object, moving eyes effectively, coordinating both eyes and eye-hand coordination. These developmental aspects of eye-movement control have been studied mainly by using scene viewing behavior with the use of artificial stimuli. However, there are only a few studies employed natural stimuli to reveal developmental changes in free viewing behavior \cite{h}. Most of the developmental studies has either focused on developmental changes during early stage in life (childhood) or during late stage in life (elderly) \cite{aring2007visual, ygge2005fixation, fioravanti1995saccadic, luna2004maturation}, and their are only a few investigating the age-related changes for whole life span (children, adult, and elderly) \cite{irving2006horizontal, h} .         

The studies reporting development of fixation system suggests that ability to fixate on target is  present since early in childhood but the more complex aspect of fixation system such as stability and control of fixation increases with increasing age between 4 to 15 years of age \cite{aring2007visual, ygge2005fixation}. Further, the studies about development of saccadic control system found that saccade velocity increases during childhood with peak value between 10-15 years of age group and then followed by decay until 86 years of age \cite{irving2006horizontal, fioravanti1995saccadic}. Findings about saccade latency i.e. reaction time to initiate an eye-movement have showed that the voluntary eye-movement decreases exponentially from birth to 14-15 years of age \cite {fischer1997development, fukushima2006vestibular, klein2001developmental, luna2004maturation}. A study \cite{fischer1997development} about anti-saccade (AS) task (which helps to understand the developmental changes in the cognitive ability to inhibit the reflexive saccade) in 300 participants of 8-65 years age has observed strong developmental effect in AS performance. The  participants of age 40-65 years of age demonstrated a moderate deterioration in AS performance. These results are also replicated in several other studies where they found the relationship between age and anti-saccade is curvliner from childhood to adulthood \cite{klein2005independent, luna2004maturation} and from adulthood to elderly stage it has been found to be linear \cite{nelson2000functional, fukushima2000development, klein2005independent}. Even though this selection of the developmental literature briefly discuss the age-impact on eye-movement control but the neglected aspect of these works is that these findings are gathered by using artificial stimulus.     

There are only a few studies employed the naturalsitic stimuls to understand the age-impact on scene viewing behavior \cite{h}. The study \cite{h} found that bottom-up features of a scene such as color, luminance, contrast, etc. guided viewing during early stage in life (7-9 years) whereas in later stage (more than 72) dominated by more top-down processing. This result can be helpful in upgrading those saliency models based on guided search theory \cite{wolfe1989guided, cave1990modeling} to make them age-adapted, by tuning the bottom-up and top-down maps according to the oberver's age-group, before combining them into a final saliency map. In this study, total number of fixation landed, inter-fixation distance, and performance in patch matching task has been used as other metrics to quantify the age-impact over scene viewing behavior. %Another study \cite{i} over naturalistic stimulus has examined the developmental changes in gaze behavior in four different group of children (2 years, 4-6 years, 6-8 years, 10 years) and adults. Similar to the previous study age-impact on saccade amplitude, fixation duration and the relation between between these fixation and saccade (ambient and focal) has been analyzed. The result revealed that fixation duration decreases while saccade length increases with age. 

%Krishna et al. \cite{am111} have parameterized the age-related differences in scene-viewing behavior for four different age groups (4 years, 6 years, 8 years, and adults) during the viewing of scenes taken from children's books and movies. They reported that explorativeness monotonically increases with age (4 years to adults), whereas the center bias is highest among younger kids (4 and 6 years). %They also proposed an age-adapted framework that can be used to predict the gaze landings of children and adult observers. 
%We extended their study and found differences in the scene-viewing behavior of children, adults, and elderly observers during inspection of natural scenes. Further, we also observed age-related differences in depth bias tendency, i.e., observers in the different age groups had different viewing tendencies for objects at different depths in the scene. 

The developmental literature presents several evidences that the age-related difference in natural scene viewing behavior is significant across children, adult, and elderly participants. Further, these findings motivates us to investigate the age-related differences in scene viewing with aim of reflecting them in development of an age-adapted computational model of saliency prediction. In order to develop and age-adapted saliency model it gets important to understand the age-impact on gaze distribution (fixation locations distribution) across different age groups,  however, the studies reported in developmental literature has mainly focused on revealing the age-impact by analyzing the gaze attributes such as fixation duration, saccade amplitude, blinks, etc. 

\subsection{Saliency Models}  

%The second part of our study is to propose an age-adapted saliency model based on the analysis result over eye-gaze data collected during scene viewing across children, adults and elderly observers. In consideration of that we present a brief literature review of computational models of visual attention system i.e. saliency modle.
The computational models of saliency prediction which are reviewed in this section have in common that they are build on the psychophysical theories of visual attention system. Treisman's Feature Integration Theory (FIT) \cite{am5} and Wolfe's Guided Search Model \cite{wolfe1989guided} has been among the most influential theories of psychological attention system. The main idea of the computational systems based on these theories is to compute different features maps in parallel and then fuse them together to get the final saliency map. 

Most of the Bottom-up feature based saliency models have a very similar general structure to compute saliency. In these models we observe following basic structure: (a) Bottom-up features such as color, intensity, and orientation are extracted over multiple scales. (b) All these features are processed in parallel, to obtain the feature maps. (c) Finally, these features are integrated to obtain the final saliency map.  Itti et al.'s model \cite{am2} is one of the most well-known model based on this general structure of saliency prediction. The Graph Based Visual Saliency Model (GBVS) \cite{am4} follows the similar general structure, where the maps are represented as fully connected graph and the distribution in Markov chain is treated as the saliency map. 

Recent studies attempted to incorporate additional cues with bottom-up features such as human intention, given task, and cognitive states of mind which are also known as top-down features. Torralba et al. \cite{am23, torralba2006contextual} proposed models for attention guidance which integrates the bottom-up features with the scene-context by using a Bayesian framework. Similarly, the SUN model of saliency prediction \cite{am17} combines bottom-up feature maps with top-down information represented as Difference of Gaussian (DoG) and Independent Component Analysis (ICA) in order to predict the salient locations in imgaes. In very recent studies supervised learning based saliency models using eye-tracking data collected over young-adults are getting popular. Judd et al. \cite{am18} proposed a model which learn to predict saliency by learning feature (bottom-up and top-down) weights in a supervised manner over 1003 images viewed by 15 young adults. In light of the recent popularity of learning based saliency models there are many eye-gaze data introduced as shown in Table 1.  It can be clearly seen form the Table 1 that participants of these experiments were mostly adults across all the dataset. Thus, all the computational models of visual attention learned on these dataset are inclined to reflect the scene exploration behavior of adults only.

\section{Dataset}

\subsection{Participants, stimuli, and apparatus}
The dataset used in this study was collected by A\c{c}iket al. \cite{h} and the data can be retrieved from \cite{am55}. Fifty-eight participants participated in this study, comprising children (18 participants, age-range 7 to 9 years, mean age 7.6), adults (23 participants, age-range 19 to 27 years, mean age 22.1), and the elderly (17 participants,  age-range 72 to 88 years, mean age 80.6) groups. All participatns reported normal or corrected-to-normal vision, including the elderly participants. Acik et al. \cite{h} declared that all participants or their parents signed a written consent to participate in the experiment. The experiment was conducted in compliance with the Declaration of Helsinki as well as national and institutional guidelines for experiments with human participants. 

 In this study, we used 192 color images belonging to 3 different categories; ``naturals", ``man-made" and ``fractals" (64 in each categories). ``Naturals" image category is representing natural scenes having trees, flowers, and bushes, this image category did not contain any artificial objects. ``Man-made" category includes urban scenes such as street, road, building and construction sites, and the ``fractals" are the computer generated shapes taken from the different web database such as Elena's Fractal Gallery. Stimulus randomizations were balanced across pairs of participants. All images had a resolution of $1280\times960$. EyeLink 1000 was used to record the gaze in its remote and hand-free mode.  
    
\subsection{Eye movement recording} 
Eye gaze was recorded while the observers viewed the images displayed for 5 seconds. The scene was subsequently replaced by a circular patch and participants had to determine if the patch was part of the previous scene or not. The patch recognition task was included to maintain the motivation of participants; thus the recognition results were not used in this study. Target stickers were placed on each observer's head to compensate for head movements. Observers viewed the stimuli from a distance of 65 cm on a 20-inch LCD monitor display (width: 40 cm). All observers were instructed to explore the scene. Fixation and saccade were identified via a fixation detection algorithm supplied by EyeLink. 

\subsection{Data representation}  
%Representing the eye-tracking data is an important step to analyze and visualize the inter-individual differences in gaze landings.
We generated human fixation maps, human saliency maps and heat maps from the gaze data. For each age group the human fixation map for an image stimuls was generated by combining the fixation landings of all observers of that age group. Further, the human saliency maps were generated from human fixation maps by convolving a Gaussian, similar to Velichkovsky et al. \cite{am13}. The heat map was obtained from the human saliency maps to visualize the age differences in the region of interest. %The process of generating these maps is illustrated in Figure 2. 

%\begin{figure*}[htp]
%  \centering
% \includegraphics[width=\textwidth, height=5cm, keepaspectratio]{DataRep}%
% \caption{The process of generating human fixation map, saliency map, and heat map}
% \end{figure*} 

\section{Analysis Method}
In this section, we discuss various measures developed for quantitative analysis of age-related changes in scene-viewing behavior of images belonging to ``naturals", ``man-made" and ``fractals" categories. Human saliency maps across age groups served as a basis for all the metrics developed for this purpose. We also propose a framework that measures age-impact over the depth bias tendency, explorativeness, inter-individual similarity, and center bias tendency for images belonging to the different categories. The details of each measure are presented in the following subsections. Fixation location was selected as prime attribute in this study to quantify the age-impact on scene viewing behavior. The reason for this selection was based on the fact that the end goal of analysis was to develop a saliency model which can predict gaze locations accurately for different age groups.

\subsection{Depth bias} 
The human visual system has the tendency of focusing earlier on the objects placed in the foreground than the objects in the background of the scene \cite{am14}. Gautier and Le Meur \cite{am15} investigated the influence of disparity on saliency for 3D conditions. Their results indicate that foreground features play an important role in attention deployment. Itti and Koch \cite{itti2001computational} suggested that stereo disparity could be used as additional cue in saliency detection.  However, this tendency of the human visual system was mainly explored for three-dimensional media but there is no study investigating the role of depth bias for two-dimensional media. Most importantly, to the best of our knowledge, for the first time we reported the role of scene depth in gaze landings across different age groups.   

 In the presented study, we visualized the age-related differences in gaze landings to the most salient foreground and the background regions in the scene. However, it is important to develop a framework to quantify this tendency across age groups. This tendency was observed in all those images in which some interesting item was placed in the foreground and the background of the scene, there was 20 such images in man-made category. 

%\begin{figure*}[!ht]
%  \centering
% 
% \subcaptionbox{\label{fig3:a} Children }{\includegraphics[width=2.2cm,height=2.2cm,keepaspectratio]{cmC}}\hspace{1em}%
% \subcaptionbox{\label{fig3:a} Adults}{\includegraphics[width=2.2cm,height=2.2cm,keepaspectratio]{cmY}}\hspace{1em}
% \subcaptionbox{\label{fig3:a} Elderly}{\includegraphics[width=2.2cm,height=2.2cm,keepaspectratio]{cmO}}\hspace{1em}
% %\subcaptionbox{\label{fig3:a}}{\includegraphics[width=1.6in]{main93}}\hspace{1em}%
% \caption{The heat map of the center-map overlay on a sample image for different age groups}
%\end{figure*}

We propose a novel framework to quantify the depth bias tendency across age groups (Figure 3). As a first step a combined saliency map for each image is generated by linearly integrating the human saliency map of children, adults, and the elderly groups. The combined saliency map is generated to find the most salient locations independent of an observer's age group. In the second step, we threshold the combined saliency map to get $t_1$ and $t_2$ percentage of the most salient locations of the scene. We fix $t_1$ and $t_2$ to five and ten, respectively, as we discovered that a small percentage of the most salient locations is sufficient to represent most of the fixation landings in the scene. In the third step, the most salient locations of the thresholded maps are manually labelled to foreground and background based on the near and far places in the scene. As shown in Figure 3, the thresholded map gives the most salient regions of the image (four different regions). Further, these regions are labelled as "F" or "B" according to their respective locations in the foreground or background of the scene (as shown in the ML map in Figure 3). Finally, the percentage of the fixations landing on the most salient foreground and background locations is calculated for each image to obtain an average score for individual age groups from the two thresholded maps. In summary, to measure the depth bias across age groups, we labelled most salient locations as foreground and background and then calculated average of the percentage of fixations landed on the foregrounds and backgrounds across all the images.   

 \begin{figure*}[htp]
  \centering
 \includegraphics[width=13cm, keepaspectratio]{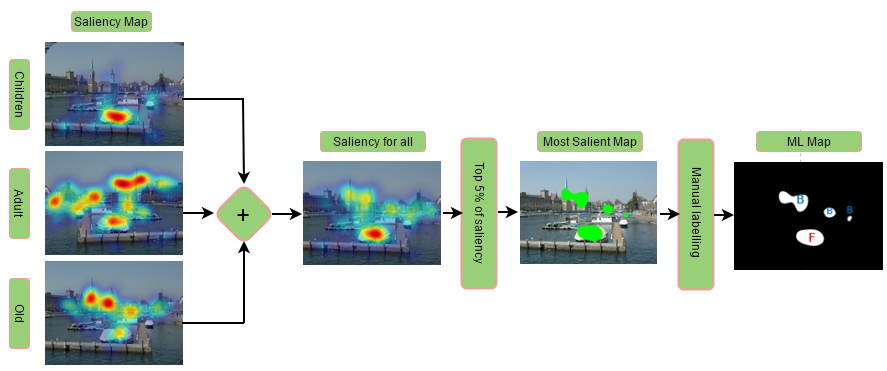}%
 \caption{Depth bias analysis framework}
 \end{figure*}

\subsection{Explorativeness}
Explorativeness study is conducted to evaluate the age-related changes in eye-movement behavior across different image categories. We found that the observers belonging to different age-groups viewing the same set of images, have differences in explored regions, and this age-impact was present independent of the image categories used in the study. We used the explorativeness metric \cite{am111} introduced previously to quantify gaze distribution of observers belonging to children, adults and elderly age groups. Explorativeness is quantified by measuring the first-order entropy of the human saliency map of an images, further, the entropy value defines the spread of fixations for different age groups. For an image, explorativeness is computed as follows:

\begin{equation}
H(U^{g}_{i})=\displaystyle\sum_{l} h_{U^{g}_{i}}(l) * \textrm{log}(L\mathbin{/}h_{U^{g}_{i}}(l))
\end{equation}  
where $U^{g}_{i}$ is the human saliency map of the $i^{th}$ image from all observers in group $g$ for which entropy is calculated and $h_{U^{g}_{i}}(l)$ is the histogram entry of intensity value $l$ in image $U^{g}_{i}$, and $L$ is the total number of pixels in $U^{g}_{i}$.  

A higher entropy correspond to more exploratory viewing behavior of the observers belonging to an age group as their gaze points are more scattered, similarly a lower entropy correspond to less exploratory behavior of the observers.   
%The explorativeness result suggests following tendency for children, adults, and elderly observers:
%\begin{itemize}

%\item Result showed significant impact of age over explorativeness tendency ($F(2,63)=179.45, p<0.001$). Post-hoc results showed that the child observers possess the lowest exploratory tendency compared to adults ($p<0.001$), who are highly explorative for the same set of images. Elderly observers showed less exploratory tendency than adults but higher than that of children ($p<0.001$) (Figure 8).  

%\item Explorativeness tendency reflects the level of details an observer visualizes in a scene. With the highest exploratory behavior, young adults have a tendency to visualize finer to coarser details in the scene, as inferred from \cite{am22}, which states that a decrease in spatial scale results in a decrease in the exploratory behavior of the observers.        
%\end{itemize}  

%The results of the explorativeness tendency for children and adults are in line with previous findings \cite{am111}. Similar results on two different class of images (\cite{h} and \cite{i}) show that this exploratory behavior is independent of the type of stimuli being observed.

%\begin{figure}[!ht]
%\centering
%\includegraphics[width=5cm,height=5cm,keepaspectratio]{explorativeness}
%\caption{Entropy across age groups}
%\end{figure}  
                           
\subsection{Inter-Individual Similarity}

Inter-individual similarity test was performed to understand the level of consistency between explored regions i.e. fixation locations among observers of the same or different age groups for all the image categories. Explorativeness analysis revealed that adults were more explorative than children and elderly participants; however, it is not clear whether the explored region by children and elderly observers was a subset of the explored region by adults (i.e. higher inter-individual similarity between children and adults age group). If that is the case, then adults can perform similarly in predicting the gaze landings of children observers to the children predicting others of the same age group. The inter-individual similarity aimed to answer this question in order to understand the need for development of an age-adapted computational attention system. We also aims to understand how the age impact differently to inter-individual similarity for the images belonging to "nature", "man-made" and "fractal" categories. 

To evaluate the inter-individual similarity of an observer's explored locations with rest of the observers of same and different age groups, we used source saliency map of one participant as a predictor to predict the target fixation of other observers (as in Figure 4). Thus under the intra-individual similarity the observer of source saliency map and target fixation points belongs to the same age group and for the inter-individual similarity the observer of source saliency map was different from target group. The formulation of this inter-individual similarity is as follows.\\

\begin{figure}[!ht]
\centering
\includegraphics[width=13cm, keepaspectratio]{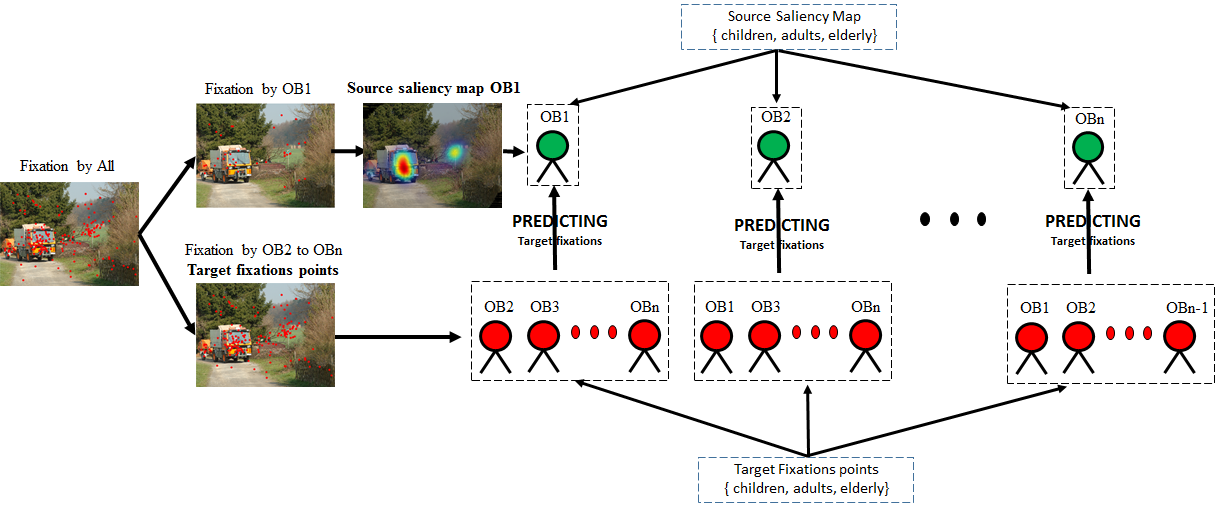}
\caption{Framework to measure inter and intra individual similarity}
\end{figure}

$A^{g_i,g_j}_{k,n}$= The prediction score of the $n^{th}$ observer of group $g_{s}$ (source saliency map) in predicting others of a group $g_t$ (target fixations) for the $k^{th}$ image.\\ 

To measure the predictibilty of an oberver's source saliency for target fixation points ($A^{g_i,g_j}_{k,n}$) we made use of area under the curve (AUC) metrics as explained in \cite{p}. The AUC-score measures how well the source saliency map of an observer could be used to find the pooled fixation locations of the rest of the observers of the same or different age group i.e. inter-individual similarity of the observer for an image with rest of observers. Final inter and intra-individual similarity score of $n^{th}$ observer is obtained by taking average over all the images ($K$) as shown in following equation. 

\begin{equation}
A^{g_i,g_j}_{n}=\frac{1}{K}\sum_{k=1}^{K}A^{g_i,g_j}_{k,n}
\end{equation}
         
Where $A^{g_i,g_j}_{n}$ is inter-individual similarity of $n^{th}$ observer, $K$ is the total number of images.  

\subsection{Center bias}
 Center bias is the tendency of fixating around the center locations in a scene while viewing the scene. This is one of the strongest bias reported in many eye tracking studies \cite{am16,am17}. The prediction accuracy of many computational models of visual attention \cite{am17,am18} have been seen to improve when center bias is included in their prediction framework. For example, a Gaussian blob centered at the middle of the image, considerably improved prediction performance of the learning-based saliency model \cite{am18}.  
  
To include the center bias tendency in the proposed age-adapted computational attention system, we first analyzed the age-related changes in this tendency. Then based on the analysis results we introduced a weighting factor for tuning the strength of the center-map in our proposed age-adapted model. To measure the center bias across age groups, we, first, generated a center-map for each age group by taking the average of all the maps for each age group. The heat map representation of the center-map over a sample image. Finally, the center bias for each age group is measured by measuring the euclidean distance between the centroid of the center-map and the center pixel of the image. %The highest euclidean distance for adults suggested the lowest center bias in adults among the age groups (203, 267, and 244 are the euclidean distance in pixels for children, adult and elderly participants).  

\section{Analysis Results}
\subsection{Depth Bias}

The results are shown in Figure 5, which show that: 
\begin{itemize}
\item Average of the fixations percentage landed on the most salient foreground locations in the images is significantly higher for child observers than for adults and the elderly (Figure 5(a)). The one-way ANOVA analysis suggest a significant influence of age over gaze landings in the foreground regions of the scene ($F(2,19)=2.99, p<0.05$), Further, a post-hoc analysis confirms that the foreground bias tendency in children is significantly different from that of adult and elderly observers ($p<0.05$).

\begin{figure*}[!ht]
  \centering
 
 \subcaptionbox{\label{fig3:a} }{\includegraphics[width=5.5cm,height=5.5cm,keepaspectratio]{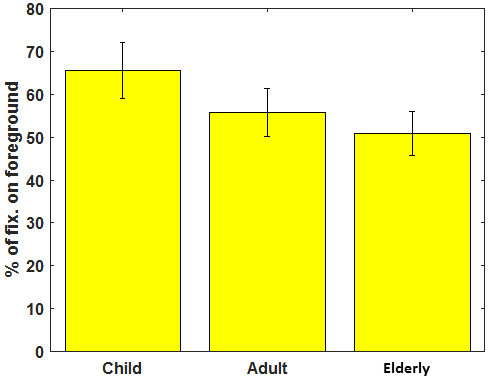}}\hspace{1em}%
 \subcaptionbox{\label{fig3:a} }{\includegraphics[width=5.5cm,height=5.5cm,keepaspectratio]{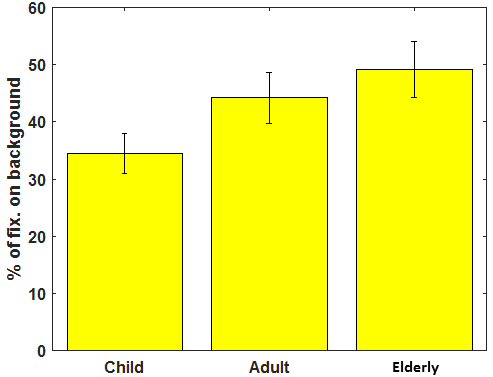}}\hspace{1em}

 \caption{(a) Percentage of the gaze landed on foreground. (b) Percentage of the gaze landed on background}
\end{figure*}

\item The percentage of fixation landings on most salient background places (Figure 5(b)) gives an interesting observation for the gaze behavior of elderly observers. It was found that elderly observers have a significantly higher tendency for fixating over items placed in the scene background than children and adult observers. The one-way ANOVA analysis confirmed the age-impact over the background bias ($F(2,19)=3.56, p<0.03$), and post-hoc analysis revealed the behavior of elderly observers is significantly different to adults ($p<0.03$) and children ($p<0.03$).
\end{itemize}

\subsection{Explorativeness}

The explorativeness result suggests following tendency for children, adults, and elderly observers:
\begin{itemize}

\item Result showed significant impact of age over explorativeness tendency across all the image categories ($F(2,191)=179.45, p<0.001$). Post-hoc results across all the image categories showed that the child observers possess the lowest exploratory tendency compared to adults ($p<0.001$), who are highly explorative for the same set of images. Elderly observers showed less exploratory tendency than adults but higher than that of children ($p<0.001$) (Figure 6).  

\item Explorativeness tendency reflects the level of details an observer visualizes in a scene. With the highest exploratory behavior, young adults have a tendency to visualize finer to coarser details in the scene, as inferred from \cite{am22}, which states that a decrease in spatial scale results in a decrease in the exploratory behavior of the observers.        
\end{itemize}  

The age-related differences in explorativeness tendency on three different class of images (see Figure 6) show that this exploratory behavior is independent of the type of stimuli being observed.

\begin{figure}[!ht]
  \centering
 
 \subcaptionbox{\label{fig3:a}}{\includegraphics[width=4.5cm,height=4.5cm,keepaspectratio]{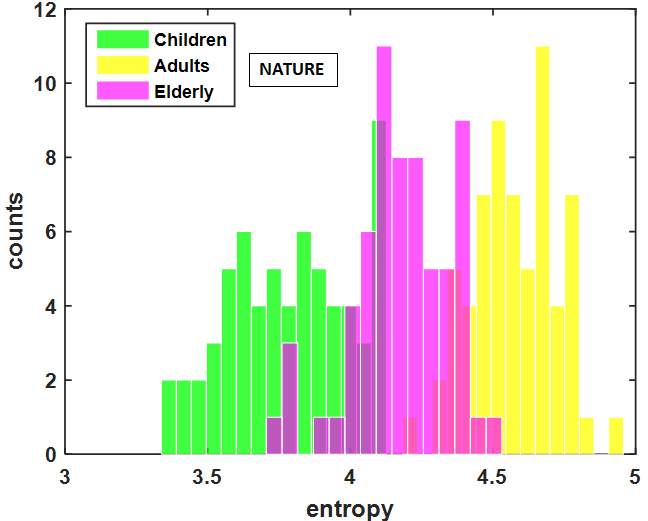}}\hspace{1em}%
 \subcaptionbox{\label{fig3:a}}{\includegraphics[width=4.5cm,height=4.5cm,keepaspectratio]{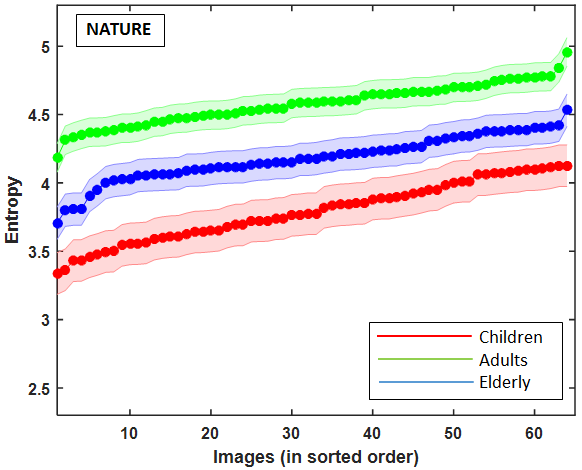}}\hspace{1em}

  \bigskip
 
 \subcaptionbox{\label{fig3:a}}{\includegraphics[width=4.5cm,height=4.5cm,keepaspectratio]{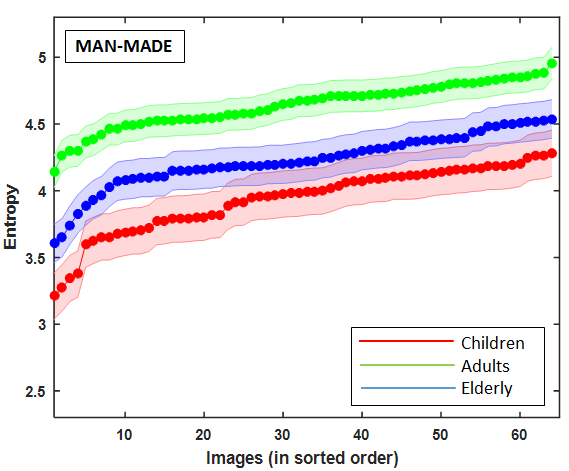}}\hspace{1em}
 \subcaptionbox{\label{fig3:a}}{\includegraphics[width=4.5cm,height=4.5cm,keepaspectratio]{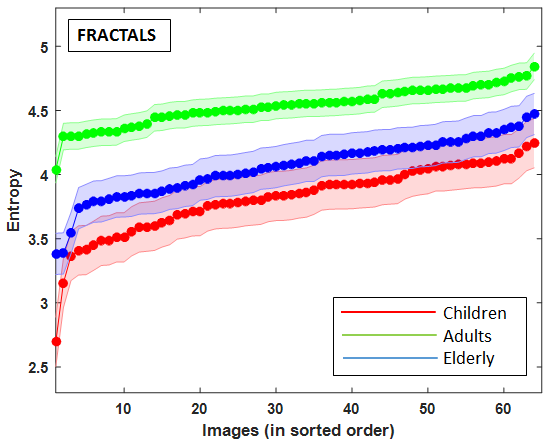}}\hspace{1em}

\caption{The age-impact over explorativeness tendency: (a) Histogram representation entropy values; showing a shift of explorativeness from left to right with increasing age (b,c,d) The figures show that age-impact over explorativeness tendency is independent of the image-type}
 
\end{figure}

\subsection{Inter-individual similarity}
The inter and intra-individual similarity results are shown in Table 1 and Figure 7. 

\begin{itemize}
\item One-way ANOVA confirmed the age influence over intra-age-group agreement between observers independent of the class of the image ($F(2,192)=41.39, \newline p<0.001$). Further, post-hoc revealed that the children are in better agreement with each other for explored locations than adults and elderly observers for all the image categories, $p<0.001$ (Figure 7(a) and Table 1).
\item Inter-age-group agreement analysis suggests that fixations of an specific age group can be best predicted by an observer of the same age group. As highlighted in table 1, the diagonal entries in each image class are highest compared to other elements in the same row, which suggest a fixation is well predicted by an observer of the same age-group than the observer of different age-group.  For instance, as shown in Figure 7(c) for natural images, a child can predict well the fixations of other children better than an adult can predict children observers ($F(2,63)=115.95, p<0.001$). 
%\item Similarly, the prediction accuracy of elderly observers (Figure 10, $A_{g_3,g_j}$) in predicting others from the same age groups was highest and significantly different from other age groups predicting elderly observers ($F(2,63)=115.95, \newline p<0.001$).  
\end{itemize}

This result suggest that the prediction accuracy of learning-based saliency models can be optimized only when the training weights are learned over the age-specific fixation data.      

\begin{figure}[!ht]
  \centering
 
 \subcaptionbox{\label{fig3:a}}{\includegraphics[width=4.5cm,height=4.5cm,keepaspectratio]{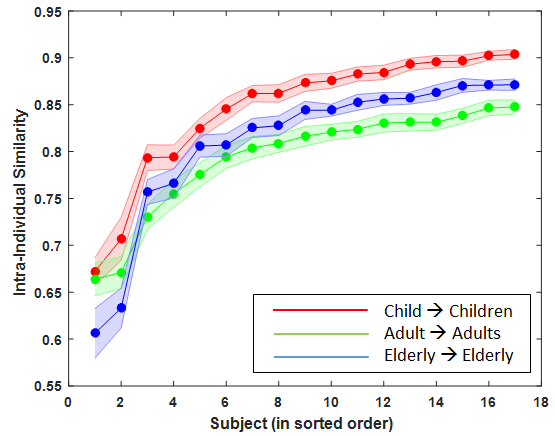}}\hspace{1em}%
 \subcaptionbox{\label{fig3:a}}{\includegraphics[width=4.5cm,height=4.5cm,keepaspectratio]{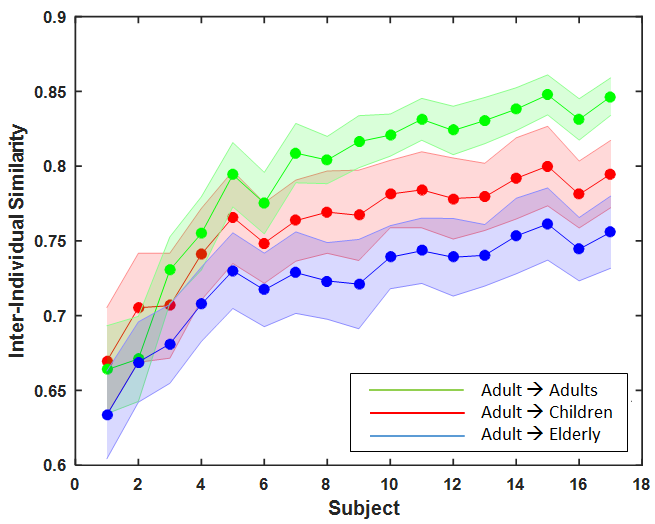}}\hspace{1em}

  \bigskip
 
 \subcaptionbox{\label{fig3:a}}{\includegraphics[width=4.5cm,height=4.5cm,keepaspectratio]{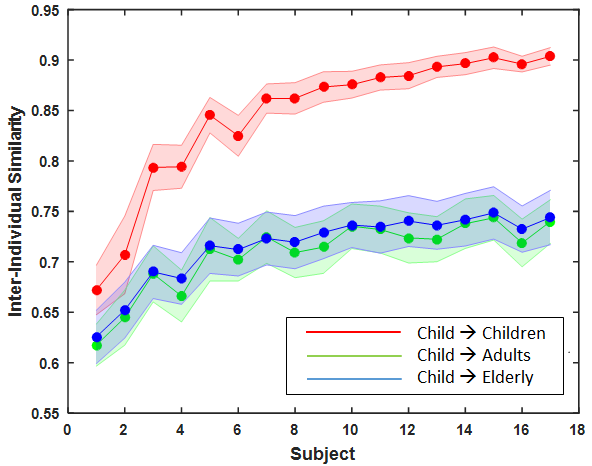}}\hspace{1em}
 \subcaptionbox{\label{fig3:a}}{\includegraphics[width=4.5cm,height=4.5cm,keepaspectratio]{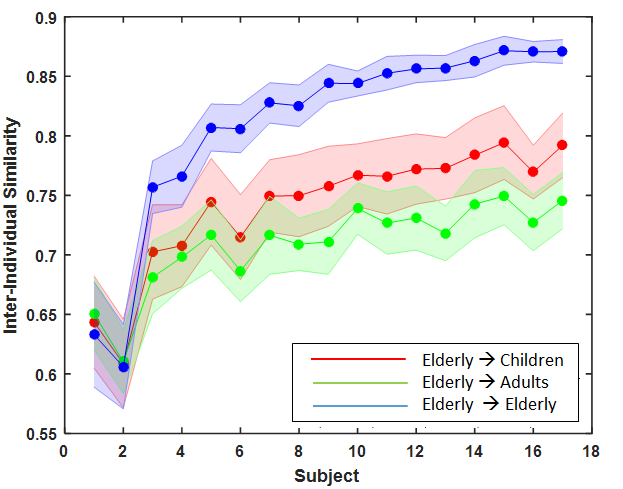}}\hspace{1em}

\caption{Inter-individual similarity: (a) Similarity score of individual participants predicting others from the same age group. (b,c,d) Similarity score of individual participants predicting others from different age group.}
 
\end{figure}

\begin{table}[]
\centering
\large
\caption{Average of the AUC score of children, adults and elderly in predicting observers of same and different age groups for different image types}
\label{my-label}
\begin{tabular}{|c|l|l|l|l|l|l|l|l|l|}
\hline
\textbf{Class}         & \multicolumn{3}{c|}{Nature}                                                                      & \multicolumn{3}{c|}{Man-made}                                                                    & \multicolumn{3}{c|}{Fractals}                                                                    \\ \hline
\multicolumn{1}{|l|}{} & \multicolumn{1}{c|}{Children}  & \multicolumn{1}{c|}{Adults}    & \multicolumn{1}{c|}{Elderly}   & \multicolumn{1}{c|}{Children}  & \multicolumn{1}{c|}{Adults}    & \multicolumn{1}{c|}{Elderly}   & \multicolumn{1}{c|}{Children}  & \multicolumn{1}{c|}{Adults}    & \multicolumn{1}{c|}{Elderly}   \\ \hline
Children               & \cellcolor[HTML]{FFCCC9}0.8390 & 0.6937                         & 0.7253                         & \cellcolor[HTML]{9AFF99}0.8483 & 0.7093                         & 0.7172                         & \cellcolor[HTML]{CBCEFB}0.8579 & 0.7257                         & 0.7559                         \\ \hline
Adults                 & 0.7548                         & \cellcolor[HTML]{FFCCC9}0.7874 & 0.7193                         & 0.7693                         & \cellcolor[HTML]{9AFF99}0.8085 & 0.7310                         & 0.7870                         & \cellcolor[HTML]{CBCEFB}0.8037 & 0.7507                         \\ \hline
Elderly                & 0.7432                         & 0.6860                         & \cellcolor[HTML]{FFCCC9}0.8063 & 0.7410                         & 0.7094                         & \cellcolor[HTML]{9AFF99}0.8093 & 0.7693                         & 0.7173                         & \cellcolor[HTML]{CBCEFB}0.8326 \\ \hline
\end{tabular}
\end{table}

\subsection{Center Bias}
We can visualize different tendency of gaze landings towards center of scene for different age groups as shown in center map of different age groups (Figure 8). Agreement score (AUC score) of fixations of different age groups towards their center map revealed the age-impact over this tendency independent of the image categories $F(2, 63)=34.10, p<0.001$ (see Figure 9). The post-hoc analysis revealed that children have significantly different tendency towards center map that adults for all the scene categories ($p<0.001$). 

Further, the result of Euclidean distance from the centriod of the explored region by different age group and center pixel of the scene showed that children have highest center bias but elderly participants showed similar tendency.  (Nature- 150, 210, and 184 are the euclidean
distance in pixels for children, adult and elderly participants, man-made- 203, 267, and 244, and fractals- 189, 214, and 202).

\begin{figure*}
\centering
  \includegraphics[width=13cm, height=5cm]{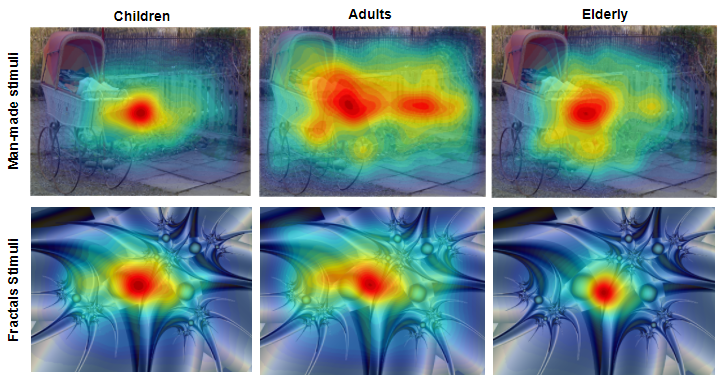}
  \caption{\textit{The heat map of the center-map overlay on a sample image for different age groups}}
  \label{fig:teaser}
\end{figure*}

\begin{figure}[!ht]
\centering
\includegraphics[width=6.5cm,height=6.5cm,keepaspectratio]{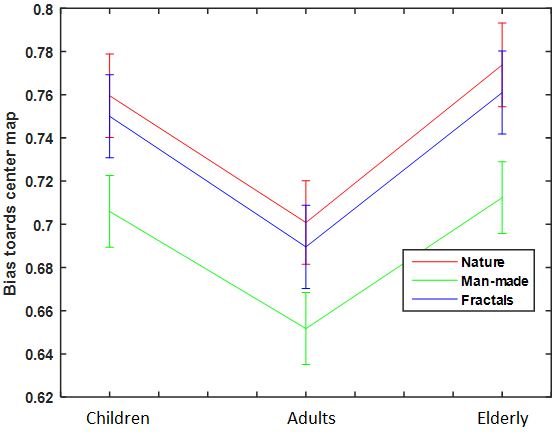}\hspace{1em}
\caption{Center bias score across age groups for all the scene categories.} 
\end{figure}

\subsection{Recommendations for age-adapted saliency model}
The analysis results recommended five major findings that we use to develop an age-adapted computational model of visual attention. These recommendations are as follows: .
\begin{itemize}
\item Depth object bias tendency showed that child observers are focused more towards the objects in foreground compared to adult and old observers. This can be incorporated in age-adapted model by including a depth map at different spatial scales for children and old age groups. 
\item  Explorativeness result showed that children were least explorative among the three age groups i.e. they are mostly focused on the coarser details of the scene, whereas being highly explorative adults evaluated finer detail too. This can be included in age-adapted model by selecting different spatial scale for different age groups. 
\item Inter age group agreement analysis suggests that the saliency maps of adults fail to appropriately predict fixation landings of child and old observers. Thus it is advisable to train the learning based models over the age-specific gaze data to optimize the prediction performance for different age groups. 
\item Analysis result of age-related changes in central bias tendency motivated us to investigate the scope of reflecting these change in computational model of visual attention by tuning the center feature map of existing algorithms. 
%\item The results of Upper Performance Limit gave us clue about the prediction accuracy of any saliency model developed for different age groups.  
       
\end{itemize} 

\section{Upper performance limit}
The final goal of our study is to reflect the age-related changes reported in the eye gaze-data analysis into the proposed age-adapted model of saliency prediction. Keeping that in mind, prior to developing any model, we have used the metrics reported in \cite{stankiewicz2011using} to estimate the upper performance limit (UPL) of an age-adapted saliency model developed for children, adults, and elderly. The UPL score also gives us an idea that how well a saliency model can predict fixations for images belonging to "natural", "man-made" and "fractals" image categories. Making it more clear, in this section we presented a robust metric UPL to measure the upper performance limit of a computational model of visual attention developed for these age groups.
 
The upper performance limit of an age group is defined as the ability of fixations over half of the observers in predicting fixations of other half of the observers. In order to do that first we generate human saliency map over fixations of first half of the observers by following the similar procedure mentioned in data representation section. Further, the human saliency map obtained after this step used to predict the fixation points of rest of the observers. The area under the curve (AUC) based metrics are used for this purpose. The framework to measure upper performance limit is shown in Figure 10.    

Higher UPL score of an age group suggests higher agreement among fixations of a group of observers with rest of the observers of same age group. Similarly the lower UPL score of an age group, suggests the lower agreement between two group of observers.     

\begin{figure}[!ht]
\centering
\includegraphics[width=13cm, keepaspectratio]{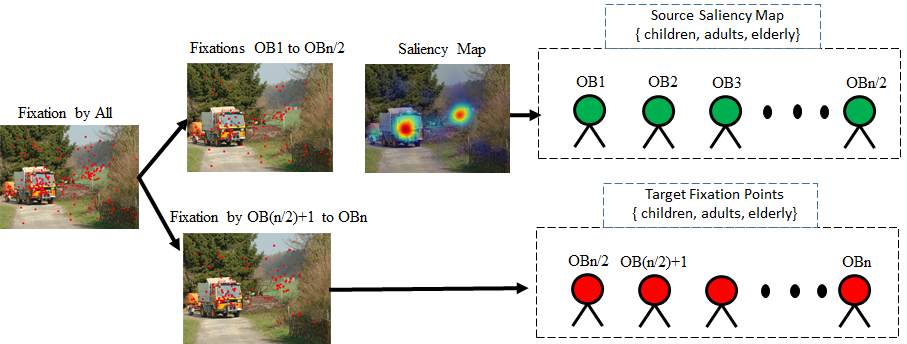}
\caption{Framework to measure upper performance limit}
\end{figure}

%\subsection{Upper Performance Limit (UPL)}

The Upper Performance Limit results are shown in Figure 11. 

\begin{itemize}
\item One-way ANOVA confirmed the age influence over UPL for all the image categories, for the man-made category $F(2, 63)=14.93, \newline p<0.001$. Further, post-hoc revealed that the UPL of children were highest and significantly different to adults and elderly participants for man-made images ($p<0.001$).     

\item Similarly, for the natural and man-made categories the UPL was significantly higher for children and elderly participants than the adults ($p<0.001$). The UPL results presents a very important finding that the prediction accuracy of a computational model developed for children will always be higher than a model developed for adults in all the image categories. %The UPL score also showed that the computational model developed for children will perform equally to the elderly.  
 
\end{itemize}

The prior knowledge about the upper-bound of the proposed age-adapted saliency model will help in optimizing the prediction accuracy for each age groups. 
  
%\begin{figure}[!ht]
%\centering
%\includegraphics[width=6cm, keepaspectratio]{UPL}
%\caption{The Upper Performance Limit (UPL) of of different age groups for natural, man-made and fractals image categories}
%\end{figure}

\begin{figure}[!ht]
  \centering
 
 \includegraphics[width=6.5cm,height=6.5cm,keepaspectratio]{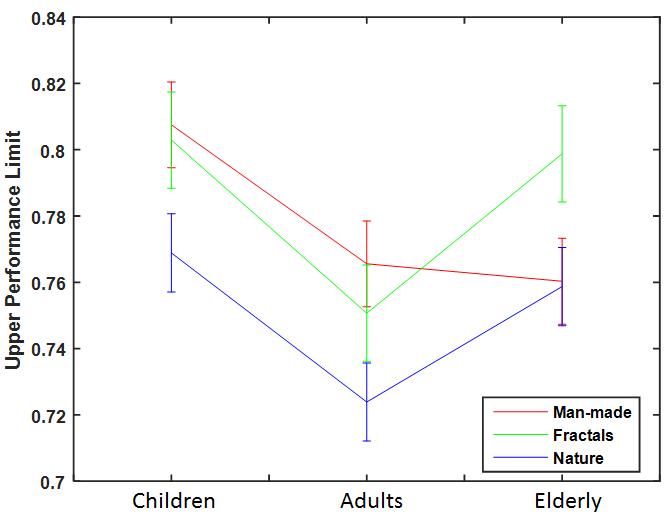}\hspace{1em}%
 \caption{ The Upper Performance Limit (UPL) of of different age groups for natural, man-made and fractals image categories.}

\end{figure}

\section{Age-adapted learning based saliency model}
In the previous section we laid some recommendations from our analysis results for developing an age-adapted computational attention system. The general structure of our proposed model is similar to learning based  saliency algorithmes  \cite{am18} in which the final saliency map  is the weighted sum of the features extracted from the input image, where the weights are learned adults gaze data. We modified learning based algorithms to reflect the age-related changes in saliency prediction following the below mentioned steps:  (a) Depth bias: We include an extra feature map for depth information at different spatial scale and orientation, generated by steerable pyramid \cite{am19}. (b)Center bias: Instead of using the center map independent of the observers age group and image category, we use a weighted mechanism for tuning them in accordance with age-related changes in center bias tendency. (c) Explorativeness: Age-related differences in explorativeness tendency was incorporated by selecting different scales of features for different age groups. (d) Individual Similarity: Finally, the model weights are learned over the age-specific gaze data instead of using adults gaze data.

\subsection{Feature used in machine learning}
We use a learning approach to train a classifier directly from age-adapted eye tracking data. The general structure is that the proposed model learns the weight for combining different features extracted from a input image to generate the final saliency map.  

Features that we used for training and testing can be categorized in low-level, mid-level, and high-level features. Low-level feature includes, color features, itti's saliency map (intensity, color, and orientation channel) \cite{am2}, Torralba's saliency map \cite{am23}, and the local energy of steerable pyramid \cite{am19}. In mid-level features horizon feature is extracted similar to \cite{am18}, as the human's have tendency of naturally directing their gaze on horizon locations in a scene viewing. For high-level features, object and car detection algorithms were applied in the proposed model. Figure 12 shows the feature maps extracted from a sample image.

The center bias is included by a adding center map generated by Gaussian blob placed at center of the scene as in \cite{am18}. The most important contribution of this work is that for the first time we included a depth feature maps extracted from input 2D scenes in context of the age-related diffrences in depth bias tendecies. The depth maps from single images are estimated using a method proposed in \cite{am20} as shown in Figure 11.

\begin{figure}[!ht]
  \centering
 
 \subcaptionbox{\label{fig3:a} Original Image}{\includegraphics[width=3cm,height=3cm,keepaspectratio]{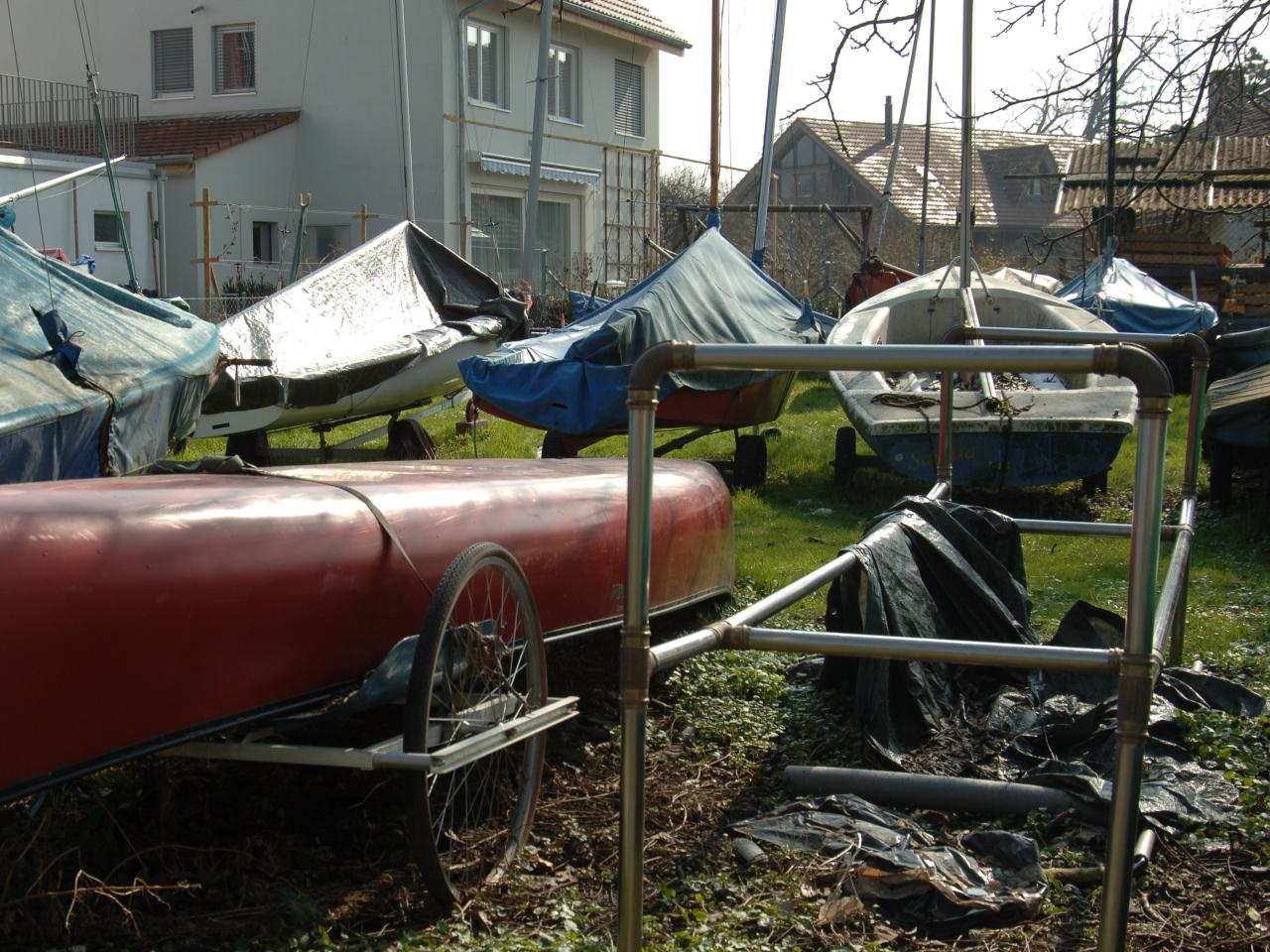}}\hspace{1em}%
 \subcaptionbox{\label{fig3:a} Depth Map}{\includegraphics[width=3cm,height=3cm,keepaspectratio]{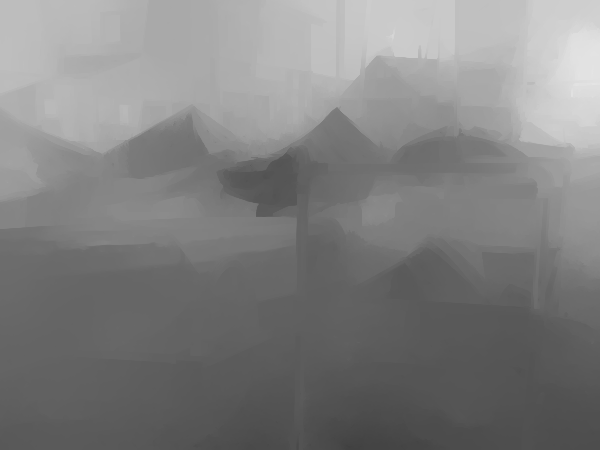}}\hspace{1em}
 \caption{Input image and its disparity map generated by the algorithm proposed by Liu et al. \cite{am20} (black area represent the nearest locations and white area is farthest locations in the scene).}
\end{figure}

\begin{figure}[!ht]
\centering
\includegraphics[width=14cm,keepaspectratio]{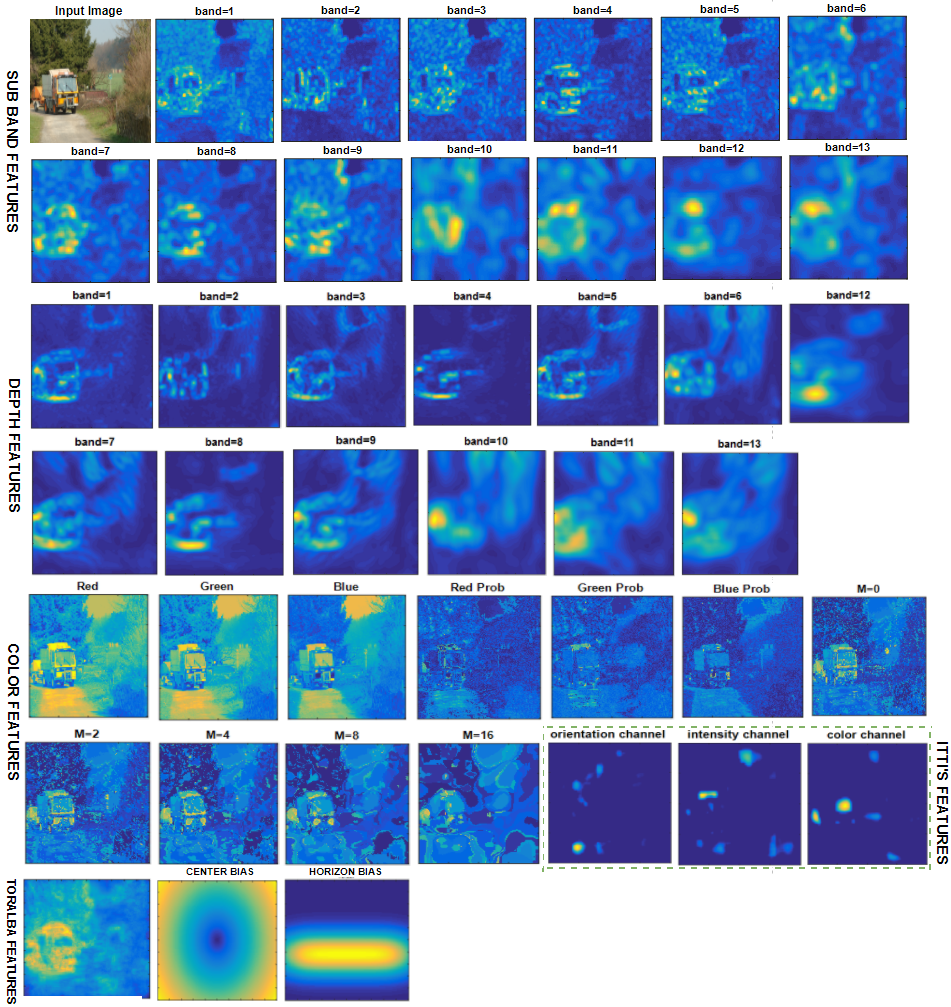}
\caption{Feature maps used in our model}
\end{figure}

\begin{table}[]
\centering
\caption{ Comparison table of our proposed age-adapted models with available computational models of saliency prediction. UPL in showing the upper performance limit of any saliency molde proposed for different age groups}
\label{my-label}
\begin{tabular}{|c|c|c|c|c|c|c|c|}
\hline
Dataset                   & Age      & Itti   & GBVS   & Patch  & Judd   & Age-adapted & UPL   \\ \hline
\multirow{3}{*}{Manmade}  & Children & 0.7140 & 0.7322 & 0.6765 & 0.7365 & 0.7456      & 0.8075 \\ \cline{2-8} 
                          & Adults   & 0.6723 & 0.6681 & 0.6083 & 0.6627 & 0.6960      & 0.7656 \\ \cline{2-8} 
                          & Elderly  & 0.6956 & 0.7160 & 0.6679 & 0.7211 & 0.7402      & 0.7603 \\ \hline
\multirow{3}{*}{Nature}   & Children & 0.6224 & 0.7427 & 0.7359 & 0.7657 & 0.7648      & 0.7689 \\ \cline{2-8} 
                          & Adults   & 0.6021 & 0.6807 & 0.6513 & 0.6774 & 0.6991      & 0.7239 \\ \cline{2-8} 
                          & Elderly  & 0.6150 & 0.7267 & 0.7186 & 0.7433 & 0.7498      & 0.7587 \\ \hline
\multirow{3}{*}{Fractals} & Children & 0.6903 & 0.7610 & 0.7111 & 0.7627 & 0.7831      & 0.8029 \\ \cline{2-8} 
                          & Adults   & 0.6582 & 0.6998 & 0.6439 & 0.6855 & 0.6960      & 0.7507 \\ \cline{2-8} 
                          & Elderly  & 0.6758 & 0.7485 & 0.7139 & 0.7562 & 0.7708      & 0.7988 \\ \hline
\end{tabular}
\end{table}

%\begin{table*}[]
%%\large
%\centering
%\caption{\textbf{Comparison table of our proposed age-adapted models with available computational models of saliency prediction.}}
%\begin{adjustbox}{max width=\textwidth}
%\begin{tabular}{|l||*{6}{c|}}\hline
%\backslashbox{Age}{Model}
%&\makebox[4.1em]{age-adapted}&\makebox[4.1em]{Judd}&\makebox[4.1em]{\text{Itti}}&\makebox[4.1em]{\text{GBVS}}&\makebox[4.1em]{\text{Patch}}\\\hline\hline
%Children & 0.7456  & 0.7365 & 0.7140 & 0.7322& 0.6765 \\\hline
%Adults & 0.6960 & 0.6627 & 0.6723 & 0.6681& 0.6083 \\\hline
%Old & 0.7402  & 0.7211` & 0.6956 &0.7160 &0.6679\\\hline
%\end{tabular}
%\end{adjustbox}
%%\textit{highest prediction accuracy for 4 year, 6 year and 8 year comes for the value of $s$ in between 2 to 4, and young achieves highest accuracy value at $s=1$}}
%\end{table*}  

\subsection{Training and Testing: age-adapted model}
We used Liblinear support vector machine to train our model over different features extracted from the scene. The model he model of Judd et al. \cite{am18} was not well suited for the age-adapted mechanism as the depth bias features maps were not included in the proposed model. To fit the training and testing in our age-adapted framework, we follow the recommendations of our analysis results. We can see in Figure 12 different features are extracted at several spatial scales (finer to coarser), where band 2 to 4 is scale 1 (finer), band 5 to 8 is scale 2 and band 9 to 12 is scale 3 (coarser).  Further, based on the results of explorativeness study, first we focus on feature scale selection, where we found that being less explorative only a few scales are required to represent the level of details explored by children and old observers, conversely, for adults, all the scales are used to represent the level of details they explored in a scene. From our experiments we found that optimal scale for children was 3, whereas scale 2 and 3 for old and scale 1, 2, and 3 for adult observers. 

We split the dataset with 192 images into 120 training and 72 test image set. Further, different features were extracted at all the scales, and only a few scales were chosen to represent the age-adapted changes in scene viewing behavior. The use of center map feature was altered by including a weight factor for tuning the age-related differences in center bias tendency as reported in our analysis. Top 10 strong positive and negative samples from top 5\% of the human saliency map and similarly 10\% strong negative pixels from bottom 20% of the map were selected selected for training.  

Further, as suggested by the agreement score analysis, strong positive and negative samples from the age-specific human saliency maps were used for training as it was seen that saliency maps of adults were poor predictors for the fixation landings of children and old observers and it is advisable to use the saliency maps of the child observers and old observers to predict fixation landings of observers of these individual groups respectively.        

For a given set of feature and levels (positive and negative samples) selected over the age-based human saliency map, SVM is used to learn the optimal weights for depth features i.e. model parameters for each age group in order to predict salient locations for different age groups. Further, for a given image of test dataset, a final saliency map is generated by combining different feature extracted at different scales over an age-specific optimal weights learned for newly included depth map during training the model. Although the prediction accuracy is mentioned for different image categories as shown in Table 2 but the model parameters are learned over all the images independent of the class of the image.    
 
\begin{equation}
  SM_g(I_i)=w_gF^T(I_i)+b_g
\end{equation}

  Where $w_g$ and $b_g$ are optimal weights for age group $g$, i.e., children, adults, and old separately, $F(I_i)$ is feature vector for the $i^{th}$ test image, this vector is of different size for different age group due to the proposed optimal feature scale mechanism. Based on the saliency values we classified the local pixel as salient or not.
  
  \begin{figure*}[htp]
  \centering
 \includegraphics[width=15cm, keepaspectratio]{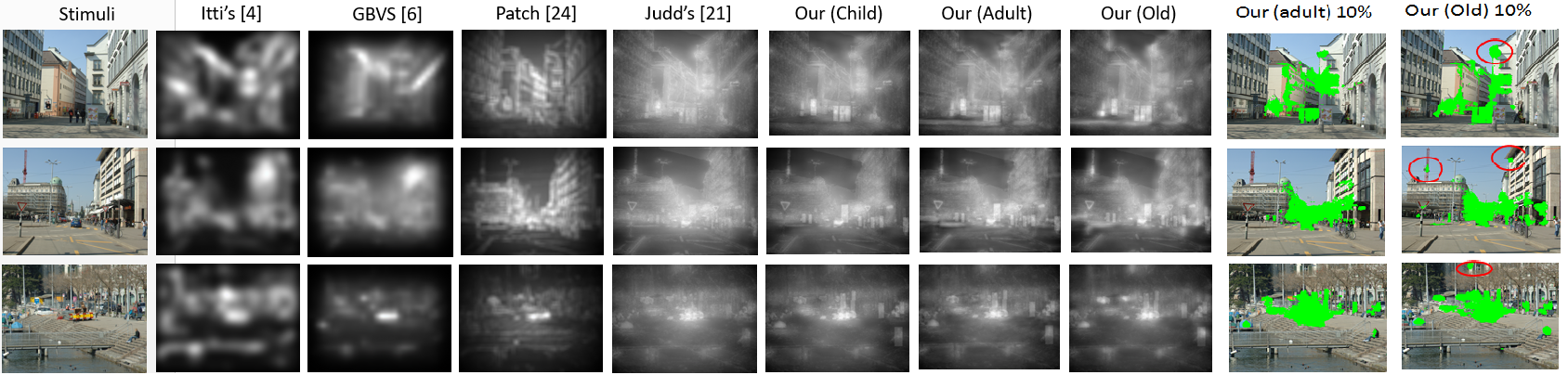}%
 \caption{Comparison of the saliency map generated from the state-of-the-art methods and and our proposed age-adapted approach, according to the hypothesis elderly focused more to the salient locations in the background and that is reflected in our proposed saliency model as shown in the red circles over the the 10\%  map.}
 \end{figure*}  
  
 Performance of our proposed age-adapted model was evaluated using AUC metrics \cite{am18}. The  evaluation score is listed in Table 2 and the resulted saliency map for sample images are shown in Figure 13. We compared our model with several state-of-the art methods reported in \cite{am2,am4,am14}. The result shows that the proposed age-adapted model outperform the existing saliency models for all age groups independent of the image categories. It is  also interesting to note that children observers show better prediction accuracy independent of the saliency model, however the age-adapted model optimized the prediction accuracy each age groups, which suggest that our modifications especially including depth map in leaning based model works well for adult observes as well. 

\section{Conclusion}   
In this work we analyzed the age-related differences in scene viewing behavior for three different age groups: children, adult and elderly observers while they viewed the scenes belonging to "natural", "man-made" and "fractals" image categories. The analysis was mainly focused to measure the age-related changes in depth bias, center bias, explorativeness, and inter-individual similarity. The result showed significant impact of age on fixation landings independent of the class of the scene viewed. Further, the analysis results helped in feature scale selection, depth map insertions, and age-specific learning in our proposed age-adapted learning based saliency model. 

The prediction accuracy of our proposed model outperforms the existing saliency models for all the age groups. The possible application of this work can be in understanding the differences in visual strategies and competencies in adults and old drivers, which can further help in optimizing the drivers training and preventing the accidents in old drivers.  Prior knowledge about scene viewing behavior of different age groups can help in designing the advertisement targeted to specific age group. 

%However, this work have limitations such as the dataset collected is only for 64 images of urban scenery, which mainly includes buildings, road, cars, etc., but the human faces, animals, and other objects are not in the current data. In future we would like to extend this work for a larger dataset and also for videos.    

\begin{acks}
We are thankful to Dr. Alpher Acik for providing the gaze data used in this study. This work is partially supported by JST CREST, JPMJCR1686, and NTT Communication Science Laboratories, NTT Corporation, Japan.
\end{acks}

% Bibliography
\bibliographystyle{ACM-Reference-Format}
\bibliography{sample-bibliography}

\end{document}